\documentclass[runningheads,a4paper]{llncs}

\usepackage{amsmath}
\usepackage{graphicx}
\usepackage[hidelinks]{hyperref}
\usepackage{url}
\urldef{\maigesis}\path|arnim.bleier@gesis.org|    
\newcommand{\keywords}[1]{\par\addvspace\baselineskip
\noindent\keywordname\enspace\ignorespaces#1}
\usepackage{tikz}
\usetikzlibrary{bayesnet}
\usepackage[numbers,sectionbib]{natbib}
\begin{document}

\mainmatter 

\title{A simple non-parametric Topic Mixture for Authors and Documents}
\author{Arnim Bleier}
\institute{GESIS - Leibniz Institute for the Social Sciences\\
Knowledge Technologies for Social Sciences\\
Unter Sachsenhausen 6-8, 50667 Cologne, Germany\\
\maigesis\\}
\maketitle

\begin{abstract}
This article reviews the Author-Topic Model and presents a new non-parametric extension based on the Hierarchical Dirichlet Process. The extension is especially suitable when no prior information about the number of components necessary is available. A blocked Gibbs sampler is described and focus put on staying as close as possible to the original model with only the minimum of theoretical and implementation overhead necessary.
\keywords{Topic models, Author models, Dirichlet process, Stick-breaking Prior, Markov chain Monte Carlo}
\end{abstract}

\section{Introduction}
Probabilistic models to infer the interests of authors have attracted much interest throughout the language modeling community, with the Author-Topic model \cite{rosen2004author} as one of its seminal representatives. Multiple modifications to the Author-Topic model have been proposed. These modifications assume either a fixed number of topics or focus on using authorship information as an additional feature in a non-parametric setting with only little resemblance to the structure of the original work. This article addresses a complementary problem -- representing the Author-Topic model in the framework of Bayesian non-parametrics but keeping as much as possible of its original structure. While this might be valuable in its own right, it is also useful in a more general sense since the steps necessary to transform an extension of Latent Dirichlet Allocation (LDA) with a fixed number of parameters to an equivalent model that grows the number of parameters with the amount data available apply to a broad range of models.

\section{Generative models for documents and authors}
We will describe two different models: The first one relates authors and documents via a fixed number of topics, and the second one models the interests of authors using a flexible number of topics. Both models are described by using the common notation of a document d being a vector of $N_d$ words, $\bf{w_d}$, where all $w_{di}$ are chosen from a vocabulary with V terms, and $\bf j_{d}$ are the authors of document d chosen from the set of all authors of size J. A corpus of D documents is then defined by the set $\{(\bf{w}_1,\bf{j}_1),..,(\bf{w}_D,\bf{j}_D)\}$.

\begin{figure}[t]
	\begin{center}
		\begin{tabular}{c@{\hspace{50pt}}r}
			\begin{tikzpicture}

  \node[obs]                              (w) {$w_{di}$};
  \node[latent, above=of w]  		  (z) {$z_{di}$};
  \node[latent, above=of z]  		  (x) {$x_{di}$};
  \node[latent, left=1.5cm of w]          (phi) {$\phi_k$};
  \node[obs, above=of x]  		  (j) {$\mathbf{j_d}$};
  \node[latent, above=of phi]  		  (beta) {$\beta$};
  \node[latent, left=1.5cm of x]          (theta) {$\theta_j$};
  \node[latent, above=of theta]  	  (alpha) {$\alpha$};

  \edge {z,phi} {w};
  \edge {x} {z}; 
  \edge {j} {x};
  \edge {beta} {phi}; 
  \edge {alpha} {theta}; 
  \edge {theta} {z};

  \plate {word} {(w)(x)(z)} {$\forall i \in [1,N_d]$} ;
  \plate {document} {(word)(w)(x)(z)(j)} {$\forall d \in \mathbf{D}$} ;
  \plate {topic} {(phi)} {$\forall k \in \mathbf{K}$} ;
  \plate {author} {(theta)} {$\forall j \in \mathbf{J}$} ;

\end{tikzpicture} &
			\begin{tikzpicture}

  \node[obs]                              (w) {$w_{di}$};
  \node[latent, above=of w]  		  (z) {$z_{di}$};
  \node[latent, above=of z]  		  (x) {$x_{di}$};
  \node[latent, left=1.5cm of w]          (phi) {$\phi_k$};
  \node[obs, above=of x]  		  (j) {$\mathbf{j_d}$};
  \node[latent, above=of phi]  		  (beta) {$\beta$};
  \node[latent, left=1.5cm of x]          (theta) {$\theta_j$};
  \node[latent, above=of theta]  	  (alpha) {$\alpha$};
  \node[latent, left=1cm of theta]  	  (tau) {$\tau$};
  \node[latent, above=of tau]  	  	  (gamma) {$\gamma$};

  \edge {z,phi} {w};
  \edge {x} {z}; 
  \edge {j} {x};
  \edge {beta} {phi}; 
  \edge {alpha} {theta}; 
  \edge {theta} {z};
  \edge {tau} {theta};
  \edge {gamma} {tau};

  \plate {word} {(w)(x)(z)} {$\forall i \in [1,N_d]$} ;
  \plate {document} {(word)(w)(x)(z)(j)} {$\forall d \in \mathbf{D}$} ;
  \plate {topic} {(phi)} {$\forall k \in [1,\infty[$} ;
  \plate {author} {(theta)} {$\forall j \in \mathbf{J}$} ;

\end{tikzpicture} 
			\tabularnewline
			a) & b)
		\end{tabular}
	\end{center}
	\caption{Admixture models for documents and authors: (a) The Author-Topic model, (b) the non-parametric Author-Topic model (this paper).}
	\label{fig:1}
\end{figure}
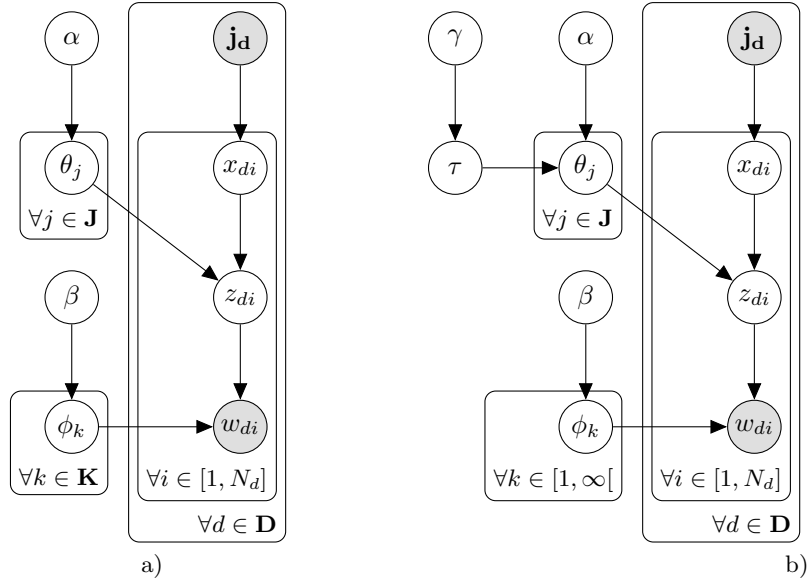

\subsection{The parametric model}
The seminal Author-Topic model \cite{rosen2004author} has two sets of unknown parameters; J distributions $\theta_j$ over topics conditioned on the authors, and K  distributions $\phi_k$ over terms conditioned on the topics - as well as the assignments of individual words to authors $x_{di}$ and topics $z_{di}$. With $\theta$ and $\phi$ being integrated out a collapsed gibbs sampler is used, analogous to \cite{griffiths2004finding}, to converge to the true underlying distributions of the Markov state variables x and z. The transitions between the states of the chain result from iteratively sampling each pair $(x_{di},z_{di})$ as a block, conditioned on all other variables:
\begin{equation}\label{gibbsLDA}
p(z_{di}=k,x_{di}=j \mid w_{di} = t, \mathbf{w}_{-di}, \mathbf{j}_d, \cdot)
\propto \dfrac{n_{jk}^{-di}+\alpha}{n_{j.}^{-di}+K\alpha}f_{k_{di}}^{-w_{di}}(t)
\end{equation}
where $n_{jk}^{-di}$ is the number of times a word of the topic k has been assigned to the author j excluding the current instance, and $\cdot$ is used in place of a variable to indicate that the sum over its values (e.g. \mbox{$n_{j.}=\sum_{k}n_{jk}$}) is taken. The assignment variable $z_{di} = k$ represents the topic of the $i^{th}$ word in document d being k as $x_{di} = j$ represents the assignment to author j. The term on the right side $f_{k_{di}}^{-w_{di}}(t)$ is the posterior density of term t under topic k:
\begin{equation}\label{wordProb}
f_{k_{di}}^{-w_{di}}(t)=\dfrac{n_{kt}^{-di}+\beta}{n_{k.}^{-di}+V\beta}
\end{equation}
where $n_{kt}^{-di}$ is the number of times a term t has been assigned to topic k again excluding the current word from the count.

\subsection{The non-parametric model}
One frequently raised question when applying the Author-Topic model to a new data set, is how to choose the number of topics \cite{sugimoto2011shifting}. The Bayesian non-parametric framework of the Hierarchical Dirichlet Process (HDP) \cite{teh2006hierarchical} offers an elegant solution to this by allowing a prior over a countably infinite number of topics of which only a few will dominate the posterior. Building on the finite version of the model we split the symmetric prior $\alpha$ over topics into a scalar precision parameter $\alpha$ and a distribution $\tau \sim Dir(\gamma/K)$. Taking this to the limit $K\to\infty$ we get the root distribution for the non-parametric Author-Topic model (fig.\ref{fig:1}b).
Analogously to the collapsed gibbs sampler for the previous LDA version we integrate over $\theta_j$, but keep $\tau$ as an auxiliary variable to preserve the structure of the state transition probabilities in the finite case for the HDP \cite{heinrichinfinite}.

\begin{equation}\label{gibbsHDP1}
p(z_{di}=k,x_{di}=j \mid w_{di} = t, {\tau}, \mathbf{j}_d, \cdot)
\propto \begin{cases} 
\dfrac{n_{jk}+\alpha{\tau}_k}{n_{j.}+\alpha}f_{k_{di}}^{-w_{di}}(t) & \text{if }  z = k \\[1em]
\dfrac{\alpha{\tau}_{k^{+1}}}{n_{j.}+\alpha}f_{k^{new}}^{-w_{di}}(t) & \text{if } z = k_{new}
\end{cases}
\end{equation}

With $f_{k^{new}}^{-w_{di}}(t)=\frac{1}{V}$ being the prior density of a word $w$ under a new topic \cite{teh2006hierarchical}. The key difference between these equations and the original model \eqref{gibbsLDA} is that we now have a root distribution ${\tau}$ for the HDP over K+1 possible states. If  there are K topics in the current step, then $\tau_{k^{+1}}$ represents the accumulated continuous probability mass of all possible but currently unused topics, allowing to choose a new one from a countably infinite pool of empty topics. If the count for number of words assigned to a topic goes to zero, the topic is returned to the pool of unused topics. 

\subsection{Sampling the Root Distribution}
However, the construction of a Markov chain for the non-parametric Author-Topic model requires that additionally the root distribution $\tau$ of the Dirichlet processes must be sampled which was not present in the finite version of the model. The discrete part of the root distribution guarantees that existing topics are reused with probability $\sum_K{\tau_k}$ and the continuous part allows for a new topic to be sampled with probability ${\tau}_{k^{+1}}$ \cite{porteous_mixture_2010}.
Given the Markov state we begin by generating J vectors
\begin{equation}
\mu_{jkr}=\dfrac{\alpha\tau_k}{r-1+\alpha\tau_k} ;\text{ with } r=1,..,n_{jk}
\end{equation}
where $n_{jk}$ are the number of words for author j which have been assigned to topic k.
Next, we draw Bernoulli random variables $m_{jkr} \sim Bern( \mu_{jkr})$. The posterior of the top-level Dirichlet process ${\tau}$ is then sampled via
\begin{equation}
{\tau} \sim Dir([m_1,..,m_k],\gamma) ;\text{ with }  m_k = \sum_{jr}m_{jkr}
\end{equation}
making $\tau$ a discrete distribution over K used topics plus one component with the probability mass of the infinite possible, yet unused topics. 

\section{Discussion}
In this work, we transformed the LDA based Author-Topic model into a non-parametric model that estimates the number of components necessary for representing the data. Yet, it will be necessary to empirically evaluate performance (i.e. perplexity) of the proposed model on benchmark data sets. While choosing the Author-Topic model as an example for such a transformation, we believe that many of the considerations made equally hold for a wider range of models and can serve as a blueprint for a simple application of non-parametric Bayesian priors.

\renewcommand\bibname{References}
\bibliography{references}{}

\end{document}